\documentclass{IEEEtran4PSCC}
\ifCLASSINFOpdf
   \usepackage[pdftex]{graphicx}
\else
   \usepackage[dvips]{graphicx}
\fi
\usepackage[cmex10]{amsmath}
\usepackage[cmex10]{amsmath}

\usepackage{cite}
\usepackage{amsmath,amssymb,amsfonts}
\usepackage{algorithmic}
\usepackage{graphicx}
\usepackage{textcomp}
\usepackage{xcolor}
\usepackage{multicol, blindtext}
\usepackage{xspace,empheq,fancybox,amsmath,amssymb,graphicx,epstopdf,epsfig,syntonly,times,amsthm} \usepackage{psfrag,color,bm,array,cite}
\usepackage{url,cite,footnote,xspace,syntonly,algorithm,algorithmic}
\usepackage{verbatim,multirow}
\usepackage[T1]{fontenc}
\usepackage{mathtools}
\usepackage{amsmath,amsfonts,amsthm,bm}
\usepackage{graphicx}
\usepackage{mwe}
\usepackage{caption}
\usepackage{stfloats}
\usepackage{amsfonts}
\usepackage{stackengine}
\usepackage{lipsum}
\usepackage{subfig}

\def\BibTeX{{\rm B\kern-.05em{\sc i\kern-.025em b}\kern-.08em
    T\kern-.1667em\lower.7ex\hbox{E}\kern-.125emX}}
    
\newtheorem{remark}{\bfseries Remark}

\newcolumntype{P}[1]{>{\centering\arraybackslash}p{#1}}

\definecolor{color_yuqi}{RGB}{39, 155, 228}

\makeatletter
\let\old@ps@headings\ps@headings
\let\old@ps@IEEEtitlepagestyle\ps@IEEEtitlepagestyle
\def\psccfooter#1{%
    \def\ps@headings{%
        \old@ps@headings%
        \def\@oddfoot{\strut\hfill#1\hfill\strut}%
        \def\@evenfoot{\strut\hfill#1\hfill\strut}%
    }%
    \def\ps@IEEEtitlepagestyle{%
        \old@ps@IEEEtitlepagestyle%
        \def\@oddfoot{\strut\hfill#1\hfill\strut}%
        \def\@evenfoot{\strut\hfill#1\hfill\strut}%
    }%
    \ps@headings%
}
\makeatother

\begin{document}
%
\title{Appliance Level Short-term Load Forecasting\\ via Recurrent Neural Network}

\author{
\IEEEauthorblockN{Yuqi Zhou$^{*}$, Arun Sukumaran Nair$^{\dag}$, David Ganger$^{\dag}$, Abhinandan Tripathi$^{\ddag}$, Chaitanya Baone$^{\dag}$, and Hao Zhu$^{*}$}
\IEEEauthorblockA{$^{*}$The University of Texas at Austin, Austin, Texas, USA\\
$^{\dag}$Eaton Research Lab, Denver, Colorado, USA\\
$^{\ddag}$Eaton Corporation, Pune, Maharashtra, India\\}
}


\maketitle

\begin{abstract}
Accurate load forecasting is critical for electricity market operations and other  real-time decision-making tasks in power systems. This paper considers the short-term load forecasting (STLF) problem for residential customers within a community. Existing STLF work mainly focuses on forecasting the aggregated load for either a feeder system or a single customer, but few efforts have been made on forecasting the load at individual appliance level. In this work, we present an STLF algorithm for efficiently predicting the power consumption of individual electrical appliances. The proposed method builds upon a powerful recurrent neural network (RNN) architecture in deep learning, termed as long short-term memory (LSTM). 
As each appliance has uniquely repetitive consumption patterns, the patterns of prediction error will be tracked such that past prediction errors can be used for improving the final prediction performance. Numerical tests on real-world load datasets demonstrate the improvement of the proposed method over existing LSTM-based method and other benchmark approaches.
\end{abstract}

\begin{IEEEkeywords}
Short-term load forecasting, recurrent neural network, long short-term memory network, deep learning.
\end{IEEEkeywords}

\thanksto{\protect\rule{0pt}{0mm} 
This work was mainly done while Yuqi Zhou was working with Eaton Research Labs.
The work of Chaitanya Baone was done while he was an employee of Eaton Research Labs.
}

\section{Introduction}
Electricity load forecasting is important for effective power system planning and operations. The long-term load forecasting problem plays an important role in system expansion planning, while mid-term and short-term load forecasting is critical for real-time operations and electricity market functions.
In this work, we will focus on the short-term load forecasting (STLF) problem, which aims to estimate the residential load for the next hour up to one week ahead.

Various machine learning methods have been proposed for effectively solving the STLF problem. These include autoregressive integrated moving average (ARIMA) model \cite{lee2011short}, support vector machine (SVM) \cite{zhang2005short}, random forest (RF) \cite{dudek2015short}, gradient boosting (GB) \cite{papadopoulos2015short}, fuzzy time series \cite{sadaei2017short}, etc. More recently, neural network (NN) based deep learning methods have been advocated for improved forecasting performance. For instance,  \cite{baliyan2015review} has presented an overview of various NN methods for the load forecasting problem, while convolutional neural network (CNN) has been recently incorporated into load forecasting for better performance \cite{sadaei2019short}.
Furthermore, the residual neural network (ResNet) model \cite{chen2018short} has also been explored for the STLF task.
Notably, the recurrent neural network (RNN) based models \cite{vermaak1998recurrent,kong2017short} have been recognized as the most powerful tool for STLF, thanks to its ability of identifying temporal patterns and correlations in time series data. 
Nonetheless, a majority of existing STLF approaches consider forecasting the aggregated demand at the system or household level, but rarely investigate  appliance-level load forecasting.
Recent advances in metering technologies have opened up the opportunity for load monitoring and forecasting at appliance level. For utility companies, aggregated appliance-level load forecasting  is important for analyzing the behaviors of residential users and  potentials of demand response. It allows for effectively designing incentive programs (e.g., \cite{xia2017energycoupon}) and encouraging the active user participation  in demand response. Residential appliance-level load forecasting has been recently investigated in \cite{din2018appliance,razghandi2020residential}. However, these proposed algorithms were developed for a very small number of households, and hence their generalizability to various types of appliances and households can come into question.

The goal of this work is to develop an efficient algorithm for the aggregated appliance-level STLF problem. Specifically, we leverage the long short-term memory network (LSTM), which is a type of RNN architecture powerful in learning long-term dependencies in time series data. 
In addition, for a time-series forecasting problem, the load forecasting error for the past data samples can become available as predictions are made. Using past forecast error, we further learn the value for the input sequences that are of similar patterns to the past ones. Thus, the  forecast error will be adaptively estimated and incorporated to improve the overall prediction performance.
The contribution of our work is three-fold:
\begin{itemize}
  \item An aggregated \emph{appliance-level} STLF algorithm is presented for residential loads using smart meter data.
  \item The proposed model can adaptively \textit{track the forecast error} from past data samples, thus improving the  prediction accuracy for each appliance.
  \item The proposed algorithm has been tested using real-world load datasets and compared with a number of popular load forecasting benchmark methods to demonstrate its performance improvement.
\end{itemize}

This paper is organized as follows. 
Section \ref{sec:intro} introduces the basics of LSTM network.
Section \ref{sec:STLF} discusses the data pre-processing step, and presents the proposed LSTM-based forecasting method which enables learning from past prediction errors. Numerical comparisons with various forecasting models are provided in Section \ref{sec:NR} to demonstrate the efficiency and error improvement of the proposed method for appliance-level load forecasting.

\section{Long Short-Term Memory (LSTM) Network} \label{sec:intro}

\begin{figure}[t!]
\centering
\includegraphics[trim=0cm 0cm 0cm 0cm,clip=true,width=0.35\textwidth]{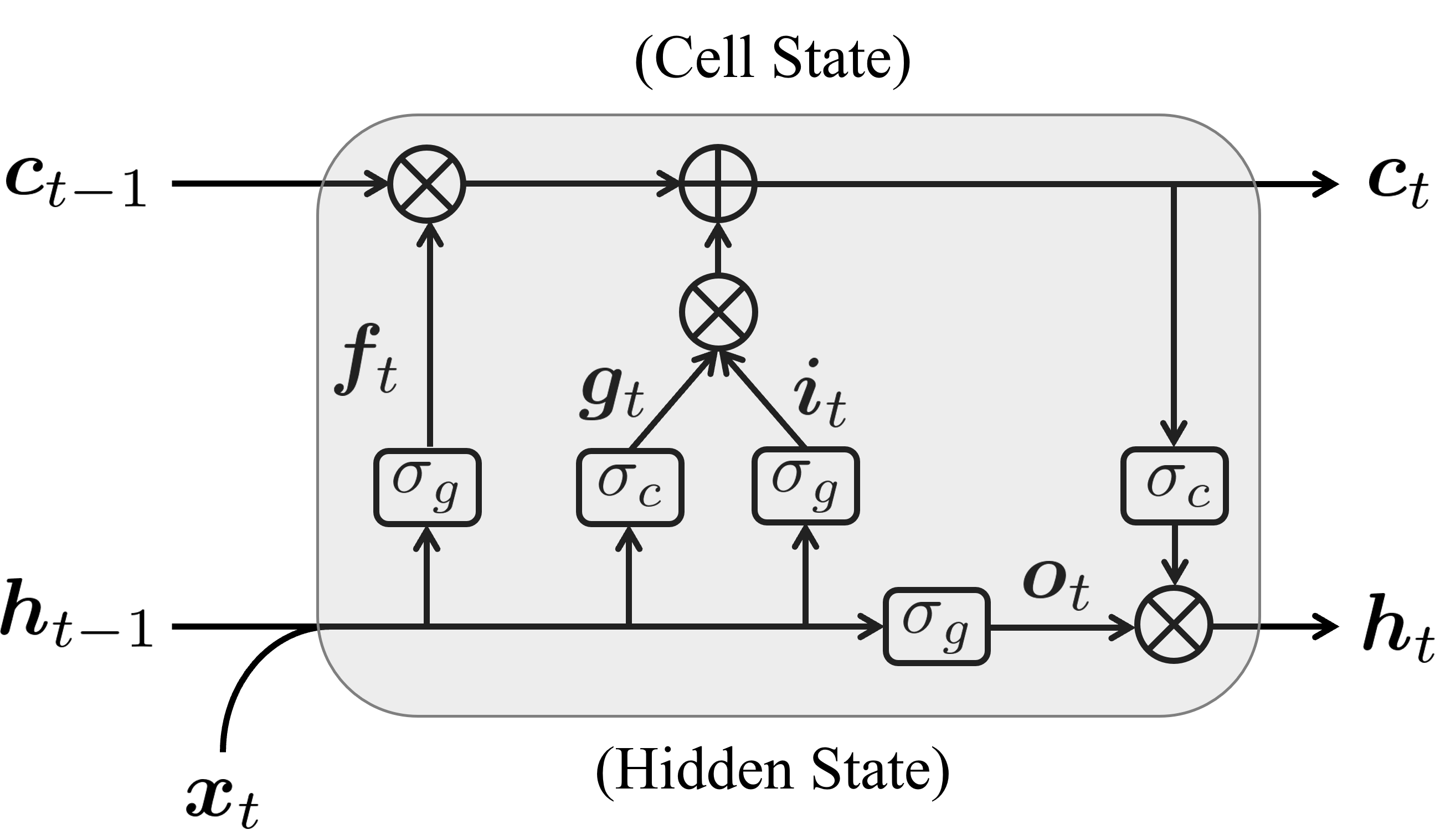}
\caption{Structure of an LSTM unit.}
\label{fig:LSTM}
\vspace*{-2mm}
\end{figure}

We first introduce the architecture of the long short-term memory (LSTM) network  \cite{hochreiter1997long}. As a special type of recurrent neural network (RNN), LSTM has been specifically designed for learning long-term dependencies in sequence data, which makes it powerful for processing time series data.  The LSTMs differ from typical RNNs mainly in the design of its repeating structure, which is termed as the LSTM unit and illustrated in Fig. \ref{fig:LSTM}. 
In standard RNNs, the repeating structure consists of simple functions such as hyperbolic tangent (tanh). In comparison, the LSTM unit includes several additional components other than the standard input $\bm{x}_{t}$ and hidden state $\bm{h}_{t}$: cell state $\bm{c}_{t}$, cell candidate $\bm{g}_{t}$, forget gate $\bm{f}_{t}$, input gate $\bm{i}_{t}$, and output gate $\bm{o}_{t}$.
The gates ($\bm{f}_{t}$, $\bm{i}_{t}$, $\bm{o}_{t}$) regulate the flow of information within the LSTM unit while $\bm{g}_{t}$ provides the candidate to be added to the cell state, as given by
\begin{subequations} \label{eq:1}
\begin{align}
  & \bm{f}_{t} = \sigma_{g} (\bm{W}_{f} \bm{x}_{t} + \bm{R}_{f} \bm{h}_{t-1} + \bm{b}_{f}) \\
  & \bm{g}_{t} = \sigma_{c} (\bm{W}_{g} \bm{x}_{t} + \bm{R}_{g} \bm{h}_{t-1} + \bm{b}_{g}) \\
  &  \bm{i}_{t} = \sigma_{g} (\bm{W}_{i} \bm{x}_{t} + \bm{R}_{i} \bm{h}_{t-1} + \bm{b}_{i}) \\
  & \bm{o}_{t} = \sigma_{g} (\bm{W}_{o} \bm{x}_{t} + \bm{R}_{o} \bm{h}_{t-1} + \bm{b}_{o})
\end{align}
\end{subequations}
where matrices $\bm{W}, \bm{R}$ and vectors $\bm{b}$ respectively denote input weights, recurrent weights and bias for each state. Furthermore, $\sigma_{g}(\cdot)$ stands for the gate activation function (sigmoid) while $\sigma_{c}(\cdot)$ the state activation function (tanh). 

The forget gate $\bm{f}_{t}$ outputs a number between $0$ and $1$ for each entry in $\bm{c}_{t-1}$, where larger values indicate the importance of keeping the corresponding values in the cell state, and vice versa. In addition,  the input gate $\bm{i}_{t}$ determines what new information to store in $\bm{c}_{t}$ by selecting from the cell candidate $\bm{g}_{t}$. Hence, the cell state keeps record of the past information in the time series and is updated as:
\begin{align}
\bm{c}_{t} = \bm{f}_{t} \circ \bm{c}_{t-1} + \bm{i}_{t} \circ \bm{g}_{t} \label{eq:2}
\end{align}
where $\circ$ denotes the Hadamard product. 
Finally, to obtain the hidden state $\bm{h}_{t}$ as the output of the LSTM unit, the updated cell state is first filtered through the tanh function ($\sigma_{c}$) and further processed by the output gate $\bm{o}_{t}$ as:
\begin{align}
\bm{h}_{t} = \bm{o}_{t} \circ \sigma_{c} (\bm{c}_{t}).  \label{eq:3}
\end{align}

In processing time series data, the current LSTM unit at time instance $t$ incorporates the hidden state $\bm{h}_{t-1}$ and cell state $\bm{c}_{t-1}$ from the previous instance, jointly with the latest input features in $\bm{x}_{t}$. The output hidden state $\bm{h}_{t}$ of the LSTM unit at time $t$ will be subsequently forwarded to a fully connected layer and a regression layer (e.g., for time series forecast) in the overall LSTM network. 
The design of gates in the LSTM network enables stable gradient propagation over time series data, which prevents the well-known gradient vanishing issues \cite{pascanu2013difficulty}.
As the cell state keeps information throughout the sequence of data processing, it perfectly resolves the issue of short-term memory that arises from the standard RNN architecture.

\section{Aggregated Appliance Level STLF}\label{sec:STLF}

This section presents the proposed STLF algorithm at the individual appliance level. Data preparation is a critical step for the implementation of machine learning algorithms in general. Thus, we first introduce the data preparation and pre-processing steps.

\subsection{Data Preparation and Pre-processing}
The availability of appliance-level consumption data is the key to the proposed method. To this end, we  use the minute-level residential household load data collected by Pecan Street. This dataset has the real power consumption that covers a wide variety of appliances (more than 50). 
For each household, we first remove data with duplicate timestamps and apply interpolation for any missing values. 
After the initial processing, there may still exist some anomaly datapoints that deviate from the typical patterns within the dataset. Thus, we have performed  a simple anomaly detection step on the updated dataset using the percentile-based approach \cite[Ch. 9]{casella2021statistical}. Basically, any values above 99th percentile are identified as outliers, and instead, replaced by the average value of previous datapoints in the time series sequence.

Certain types of appliances within the household may share similar consumption patterns. For this reason, we combine the power consumption data from homogeneous types of electric appliances (e.g., appliances in bathroom 1 and bathroom 2) and name the combined data according to their categories (e.g., bathroom). For a given electric appliance, the aggregated data across multiple co-located users is more statistically significant and suitable for forecasting as compared to that of an individual user. Thus, we aggregate power consumption data of users within a community and perform the load forecasting on the total consumption for each individual appliance. Finally, we have implemented zero-center normalization for the resultant aggregated data for improving the numerical stability. 

This pre-processed power consumption data is ready for setting up the input features and target variables of the LSTM forecasting model. The target variable at time $t$ is essentially the power consumption $p_t$ to predict, while the input feature vector $\bm{x}_{t}$ is composed of power consumption data of several previous timestamps:
\begin{align}
\bm{x}_{t} = \left[p_{t-\tau}, \cdots, p_{t-2}, p_{t-1}\right]  \label{eq:4}
\end{align}
where $\tau$ is a chosen time interval.

\subsection{LSTM with Feedforward Control}
Although LSTM is known to be very powerful in dealing with time series data, the current architecture fails to incorporate historical results to improve the forecasting performance. 
In fact, in the training stage the prediction error for every time instance is readily available upon obtaining the forecast value. As more predictions are made, more historical datasets (input sequences and the corresponding errors) become available, which can provide us with additional information to improve the further predictions.
Many previous research works only consider adding different types of hidden layers (e.g., \cite{kong2017short,razghandi2020residential}) to improve the performance, but none of them attempt to establish a learning process between the forecast error and the input features using historical data.

\begin{figure}[t!]
\centering
\includegraphics[trim=4cm 4cm 4cm 5.5cm,clip=true,width=0.5\textwidth]{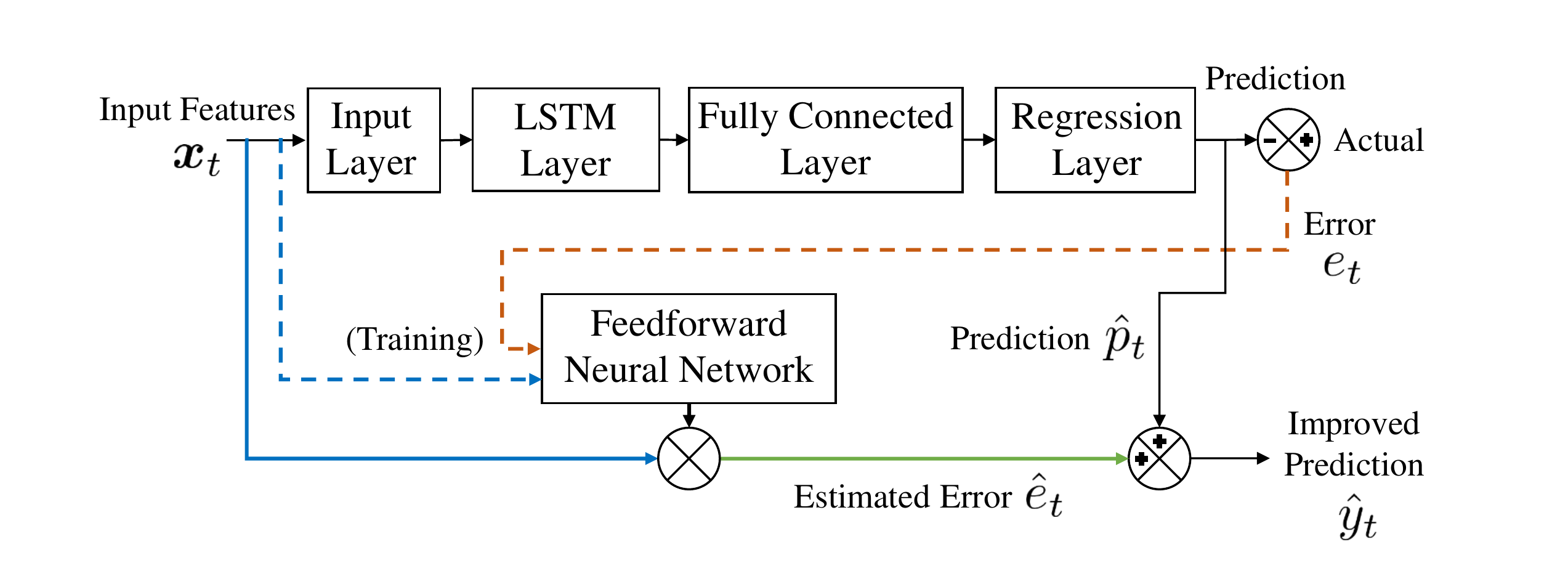}
\caption{Diagram of the augmented LSTM forecasting model featuring adaptive learning from historical repetitive patterns.}
\label{fig:structure}
\vspace*{-2mm}
\end{figure}
\begin{figure}[t!]
\centering
\includegraphics[trim=2.5cm 4.5cm 3.5cm 3.5cm,clip=true,width=0.5\textwidth]{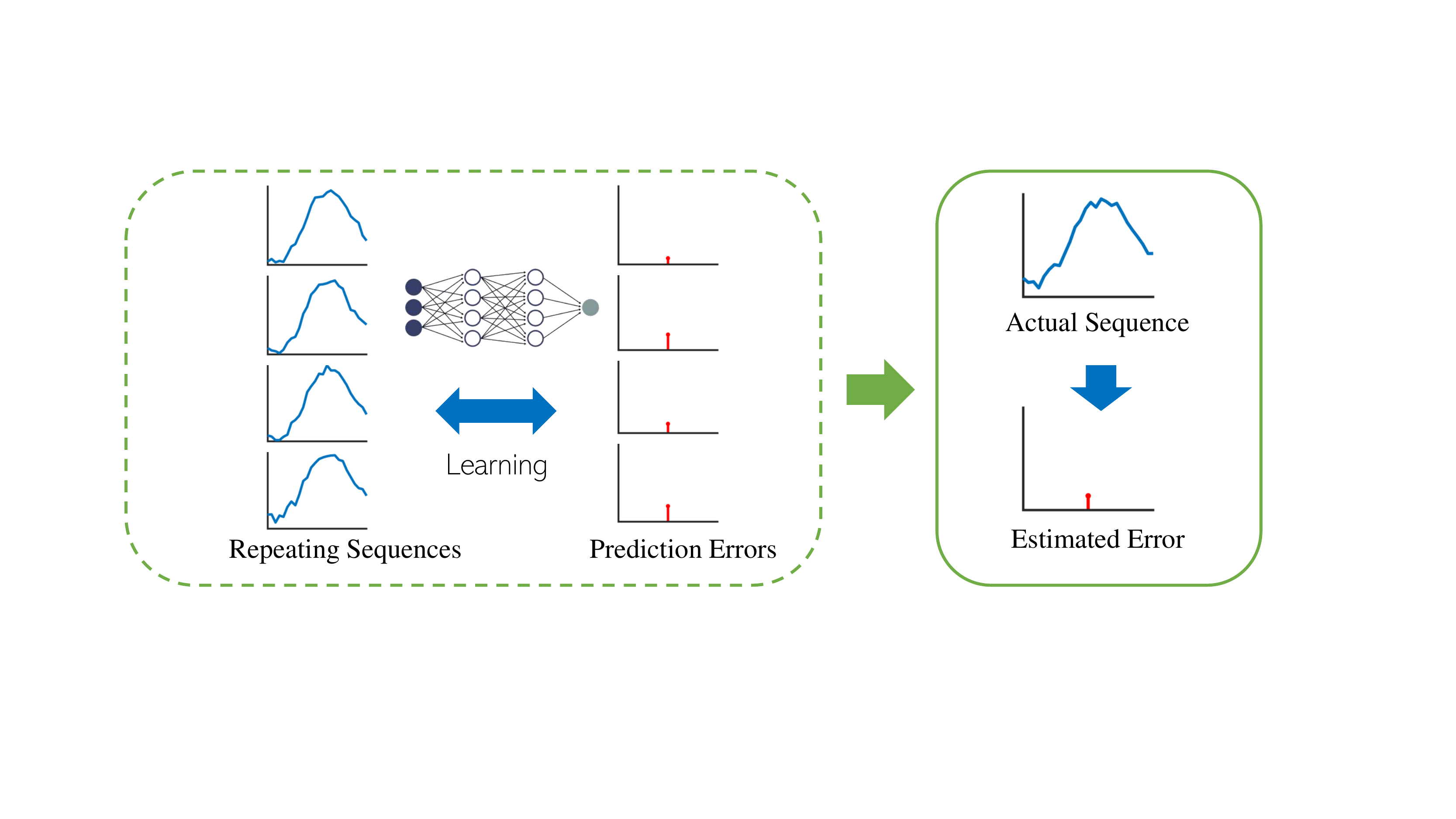}
\caption{Illustration of learning forecast error with aggregated load data for improved prediction performance.}
\label{fig:learning}
\vspace*{-2mm}
\end{figure}

To this end, we propose an augmented LSTM model which features an additional feedforward neural network that adapatively learns the connection between repetitive consumption patterns and prediction errors, which is illustrated in Fig.~\ref{fig:structure}. 
The error ${e}_{t}$ for an input feature $\bm{x}_{t}$ can be predicted using:
\begin{align}
{e}_{t} = \sigma_{p}\left(\bm{w}^{\mathsf T}_{p}\bm{x}_{t} + b_{p} \right)  \label{eq:5}
\end{align}
where $\sigma_{p}$ denotes the activation function (e.g., ReLU), $\bm{w}_{p}$ and $b_{p}$ are weights and bias for the feedforward neural network. 
Notice that this neural network is not recurrent based and is implemented separately from the standard LSTM network. Therefore, there are no recurrent weights involved as in the LSTM network.
This structure enables us to learn from previous mistakes, and the output for this neural network (predicted error) will be added to any upcoming prediction as an additional adjustment to improve the performance. 
The process of learning prediction error is illustrated in Fig. \ref{fig:structure} and Fig. \ref{fig:learning}.
Thus, the corrected prediction at time $t$ becomes:
\begin{align}
\hat{y}_{t} = \hat{p}_{t} + \hat{e}_{t}  \label{eq:6}
\end{align}
where $\hat{p}_{t}$ is the predicted power consumption and $\hat{e}_{t}$ is the estimated forecast error.


The training of this additional feedforward NN requires sufficient historical data of input sequences and prediction errors. Since the proposed framework records the original LSTM forecast errors ${e}_{t}$, the learning model can be improved progressively by adding new data once it is available.
The idea of learning prediction errors is attainable largely because the appliance-level consumption data tends to have unique recurring input sequences and errors. The feasibility is briefly explained as follows.

\begin{remark}[Feasibility of learning forecast error]
While the appliance-level power consumption data usually varies for individual users, the total consumption for one appliance over all users normally has stronger periodicity. Thus, in our prediction task, the relationship between these recurring patterns and the resulting forecast errors can be efficiently learned using historical and real-time data.
\end{remark}

The performance of LSTM on load forecasting can be considerably affected by the hyperparameter settings. Therefore, previous research works normally require hyperparameter tuning to determine the best setting. Through learning the consumption patterns for each appliance, the proposed
framework can effectively offset the inaccuracy caused by unsatisfactory hyperparameter settings. This can save numerous validation experiments before the actual forecast is made. 
Furthermore, since the learning model can be updated adaptively with new data, the performance is consistent even if the consumption patterns change over time (e.g, seasonal changes).

\section{Numerical Results} \label{sec:NR}
This section presents the numerical performance of the proposed load forecasting method together with a number of benchmark approaches.
The data cleaning and pre-processing are first completed using Python, and the RNN-based load forecasting methods are next implemented using MATLAB\textsuperscript{\textregistered} deep learning toolbox.
The simulations are performed on a regular laptop with Intel\textsuperscript{\textregistered} CPU @ 1.70 GHz and 8 GB of RAM. The Adam optimizer \cite{kingma2014adam} is used for the LSTM training, with a max of 500 epochs. We set the gradient threshold to be $0.01$ and the learning rate to be $0.0001$. The input sequence length is chosen to be $\tau = 12$ and the number of hidden units is 200. For the feedforward neural network, we use a simple two-layer shallow neural network. One can also utilize deep neural networks by adding more hidden layers. 

\begin{table*}[hbt!] 
  \centering
  \caption{Comparisons of mean absolute percentage error (MAPE) metric for different forecasting models}
\begin{tabular}{ |P{2.4cm}||P{1.3cm}|P{1.3cm}|P{1.3cm}|P{1.3cm}|P{1.3cm}|P{1.3cm}|P{1.3cm}||P{1.3cm}| }
 \hline
 & {RF} & {SVM} & {KNN} & {GB} & {BR} & {MLP} & {LSTM} & {Proposed}  \\
 \hline
\multirow{1}{8em}{AC} 
 & 87.50	& 78.55	& 93.94	& 92.89	& 102.50	& 381.43 & \textbf{41.84} & 18.89
\\
 \hline
  \multirow{1}{8em}{EV} 
  & 868.8 &	931.44 &	765.28 &	872.54 &	926.61 &	1303.4 &	\textbf{205.40} &	126.37
\\
\hline
 \multirow{1}{8em}{Bedroom}  
  & 13.80 &	16.77 &	14.03 &	13.70 &	12.34 &	17.36 &	\textbf{4.79} &	4.21
\\
 \hline
 \multirow{1}{8em}{Bathroom}  
  & 83.65 &	63.98 &	73.77 &	86.78 &	85.81 &	128.89 &	\textbf{48.74} &	41.59
\\
 \hline
  \multirow{1}{8em}{Dining Room} 
  & 59.78 &	71.82 &	48.85 &	56.20 &	65.22 &	141.88 &	\textbf{48.68} &	42.83
\\
 \hline
 \multirow{1}{8em}{Living Room}  
  & 20.13 &	36.02 &	25.25 &	19.51 &	20.18 &	22.33 &	\textbf{6.44} &	5.64
\\
 \hline
  \multirow{1}{8em}{Outside} 
  & 100.94 &	48.73 &	71.28 &	49.41 & 152.25 & 556.26 & \textbf{13.64} & 10.37
\\
 \hline
 \multirow{1}{8em}{Furnace}  
  & 74.32 &	48.65 &	69.39 &	60.96 &	58.79 &	65.30 &	\textbf{9.58} &	8.53
\\
 \hline
 \multirow{1}{8em}{Washer} 
  & 76.44 &	108.17 &	111.25 &	84.27 &	71.72 &	110.38 &	\textbf{49.36} &	40.34
\\
 \hline
  \multirow{1}{8em}{Pool} 
  & 37.21 &	48.17 &	35.79 &	34.93 &	35.80 &	45.46 &	\textbf{12.32} &	11.20
\\
 \hline
  \multirow{1}{8em}{Cooler} 
  & 22.17 &	22.21 &	23.95 &	22.61 &	21.95 &	23.00 &	\textbf{21.44} &	18.69
\\
 \hline
  \multirow{1}{8em}{Garage} 
  & 31.30 &	21.55 &	37.35 &	33.66 &	31.11 &	51.78 &	\textbf{13.05} &	11.32
\\
 \hline
  \multirow{1}{8em}{Kitchen} 
  & 18.77 &	33.78 &	22.91 &	18.24 &	19.03 &	16.23 &	\textbf{11.77} &	10.29
\\
 \hline
  \multirow{1}{8em}{Disposal} 
  & 14.86 &	82.10 &	16.10 &	15.35 &	14.87 &	23.56 &	\textbf{14.15} &	13.55
\\
 \hline
  \multirow{1}{8em}{Pump} 
  & 28.70 &	75.12 &	\textbf{14.77} &	28.33 &	15.28 &	16.90 &	15.69 &	12.41
\\
 \hline
  \multirow{1}{8em}{Lights} 
  & 11.75 &	25.11 &	12.50 &	11.80 &	11.14 &	12.19 &	\textbf{4.48} &	4.40
\\
 \hline
  \multirow{1}{8em}{Refrigerator} 
   & 9.88 &	12.82 &	11.84 &	9.67 &	9.48 &	\textbf{9.30} &	9.71 &	4.64
\\
 \hline
  \multirow{1}{8em}{Total} 
   & 25.60 &	30.28 &	28.85 &	26.53 &	26.22 &	31.88 &	\textbf{9.72} &	6.62
\\
 \hline
\end{tabular}
  \label{tab:1}
  \vspace*{-2mm}
\end{table*}

The appliance-level power consumption data is aggregated over 289 individual households located in Austin, Texas. Furthermore, the minute-level data is converted to 15-minute interval data for day-ahead load forecasting. We choose load data from January to March in 2020, in which the first two months are used for training while March is for testing.
To evaluate the performance of the proposed algorithm, we have also implemented several popular load forecasting methods apart from the standard LSTM method. These include random forest (RF), support vector machine (SVM), K-nearest neighbors (KNN), gradient boosting (GB), Bayesian ridge (BR), and multi-layer perceptron (MLP). We have picked about 25 appliances (locations) in total for simulation studies. 

The mean absolute percentage error (MAPE) can be used to quantify the forecasting accuracy, and the metric is defined as the mean of absolute error normalized over the actual data value:
\begin{align}
\textrm{MAPE} = \frac{100}{n} \sum_{i=1}^{n} \left|\frac{y_i - \hat{y}_i}{y_i}\right|.  \label{eq:7}
\end{align}
The MAPE values for load forecasting of selected appliances under different models are recorded in Table \ref{tab:1}. 
Compared with other benchmark methods, the standard LSTM has better forecasting performance for most of the listed appliances, thanks to its ability to learn long-term dependencies.
The lowest MAPE values for each appliance under all the benchmark methods (including LSTM) are marked in bold.
Under the standard LSTM model, the MAPE values for most appliances fall within the range of 10 to 30. 
The randomness of power consumption for certain appliances (e.g., electric vehicle (EV) charging, washer) are higher, which makes them more difficult to predict and results in large MAPE values. On the other hand, power consumption at certain locations (e.g., living room, bedroom) is more patterned and predictable, leading to small MAPE values. The total power consumption has relatively smaller MAPE values as compared to many appliances, as aggregating the consumption of different appliances can mitigate the randomness of data. For example, if one appliance is in use, then it is highly unlikely that certain appliances are also in use. Thus, the total consumption tends to stabilize and have fewer fluctuations.
\begin{remark}[]
The load forecasting accuracy can be notably improved by increasing the number of training samples. In addition, an increased level of customer aggregation can lead to an improved prediction performance. This observation has been reported in \cite{quilumba2014using,stephen2015incorporating}, as the aggregated demand tends to show consistent consumption patterns that facilitate the prediction process.
\end{remark}

\begin{figure}[hbt!]
\centering
    \centering
    \subfloat[AC]
    {
        \includegraphics[trim=0cm 0cm 0cm 1.2cm,clip=true,totalheight=0.17\textheight]{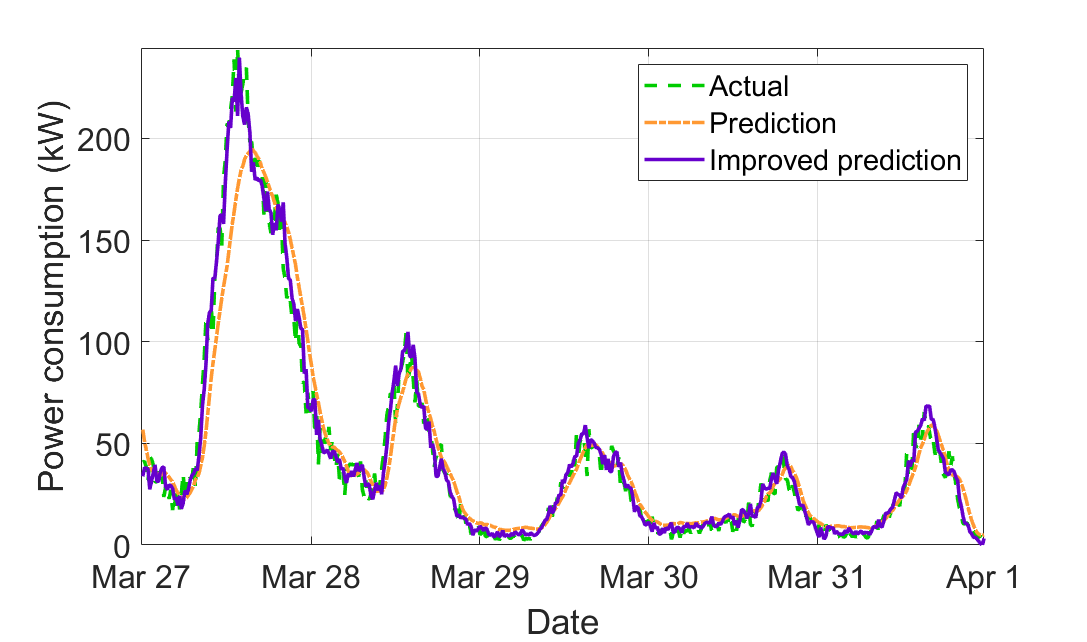}
        \label{fig:AC} 
    } \vspace*{-2mm}
    \\
    \centering
    \subfloat[EV]
    {
        \includegraphics[trim=0cm 0cm 0cm 1.2cm,clip=true,totalheight=0.17\textheight]{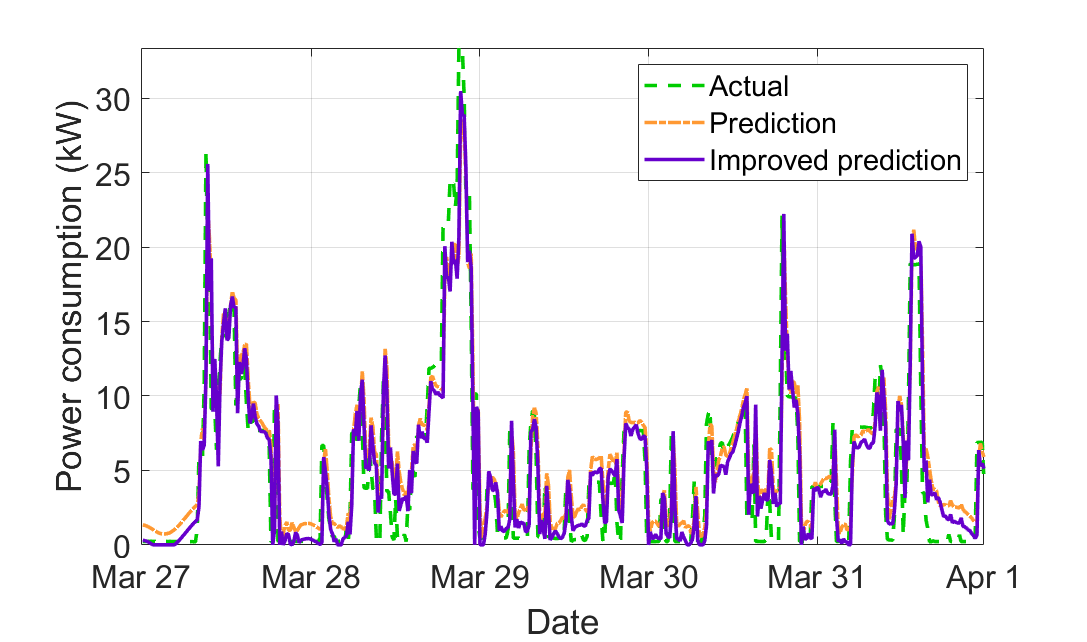}
        \label{fig:EV} 
    } \vspace*{-2mm}
     \\
    \centering
    \subfloat[Refrigerator]
    {
        \includegraphics[trim=0cm 0cm 0cm 1.2cm,clip=true,totalheight=0.17\textheight]{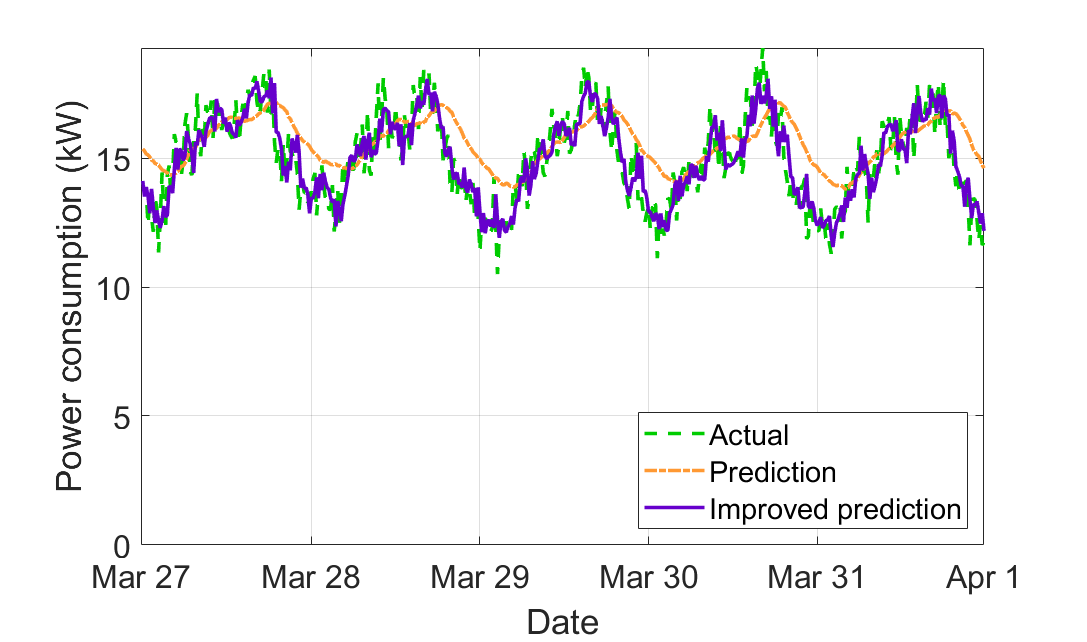}
        \label{fig:refrigerator} 
    } \vspace*{-2mm}
    \\
    \centering
    \subfloat[Total]
    {
        \includegraphics[trim=0cm 0cm 0cm 1.2cm,clip=true,totalheight=0.17\textheight]{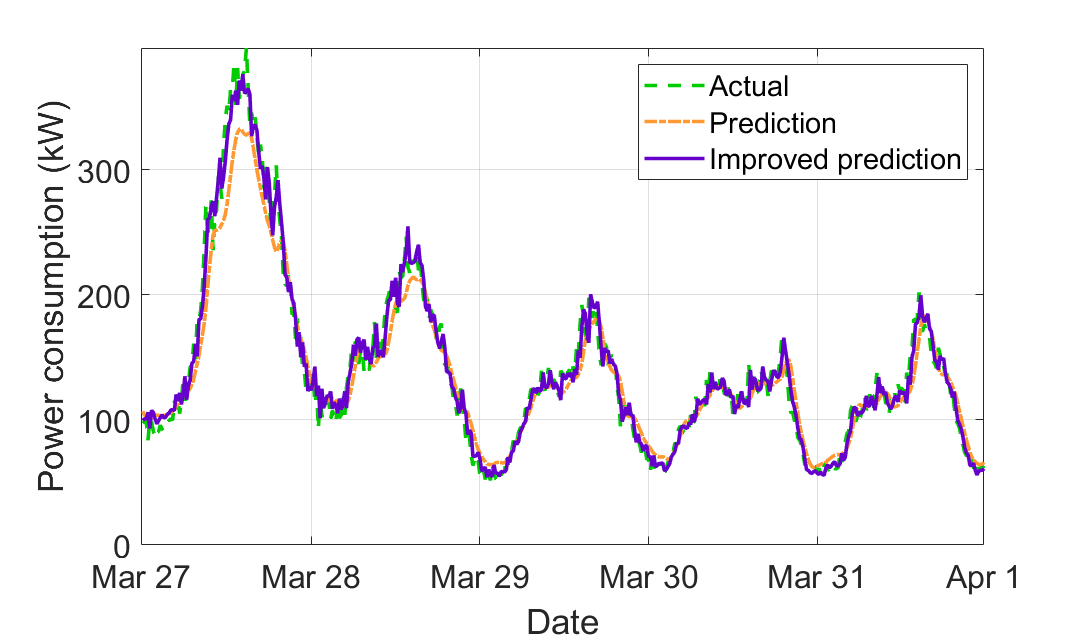}
        \label{fig:total} 
    } 
    \caption{Power consumption prediction for selected appliances.}
    \label{fig:sample_subfigures}
    \vspace*{-1mm}
\end{figure}

The standard LSTM method has an average MAPE value of 30.0, while the proposed algorithm with augmented feedforward control achieves an average MAPE value of 21.7, which reduces the error by $28\%$ on average.
The comparison of load forecasting performance for air conditioning (AC), electric vehicle (EV), refrigerator and total consumption under the two models is shown in Fig. \ref{fig:sample_subfigures}. It provides the forecast values together with the actual power consumption for 5 consecutive days in March. 
Notice that although the load forecasting for EV has high MAPE values (see Table \ref{tab:1}), the prediction results look acceptable here. This is because the EV consumption is only available for limited number of customers, therefore at given time the aggregated power consumption can be either zero or close to zero. The difficulty of precisely matching these valley points normally leads to larger MAPE values, but the overall consumption pattern is well captured.

Compared with the standard LSTM, the proposed method reduces the forecast error by $54.9\%$, $38.5\%$, $52.2\%$ and $31.9\%$ for the selected appliances, respectively. Clearly, the forecasting with error adjustment helps shift the power consumption pattern to be closer to the actual values. The AC and refrigerator have relatively higher error reduction rate, as they both have more repetitive consumption sequences compared to others. This makes the feedforward learning more efficient during the training phase, thus leading to better performance. In comparison, the EV charging consumption is more stochastic and does not show strong periodicity. Some sequences may rarely or never appear during the training phase, which makes the learning less efficient, thus offering less significant reductions. Overall, the proposed method can adaptively adjust the prediction values by learning from the historical forecasting data and provide improved forecasting performance.

\section{Conclusions and Future Work}
\label{sec:con}
In this paper, we propose a short-term load forecasting framework for effectively predicting the total power consumption of individual appliances at community level.
We deploy the LSTM network, and build in a feedforward neural network to enable learning from historical repetitive sequences to the prediction errors. Thus, each initial prediction result can be tuned by an estimated error term for better forecasting performance. Numerical tests using actual residential load data show the validity and improved performance of the proposed algorithm.
Future work will investigate the load forecasting and monitoring under limited sensors. We are also interested to explore attention-based neural networks \cite{vaswani2017attention} for the load forecasting task.



%

\bibliography{bibliography.bib}

\bibliographystyle{IEEEtran}

\itemsep2pt

\end{document}